\documentclass[letterpaper, 10 pt, conference]{ieeeconf}

\IEEEoverridecommandlockouts  

\usepackage{graphics}
\usepackage{graphicx}  
\usepackage{float}
\usepackage{amsmath}
\usepackage{amssymb}
\usepackage{mathrsfs}
\usepackage{cite}
\usepackage[normalem]{ulem}
\usepackage{color}
\usepackage{longtable,tabularx}
\usepackage{makecell}
\usepackage{multirow}

\title{\LARGE \bf
Energy-Optimal Planning of Waypoint-Based UAV Missions \\- Does Minimum Distance Mean Minimum Energy?
}
\author{Nicolas Michel $^{1}$, Ayush Patnaik $^{1}$, Zhaodan Kong $^{1}$, Xinfan Lin $^{1*}$
\thanks{$^*$This work was supported by funding from the Office of Naval Research NEPTUNE program [award number N00014-21-1-2080].}
\thanks{$^{1}$Nicolas Michel, Ayush Patnaik, Zhaodan Kong and Xinfan Lin are with the Department of Mechanical and Aerospace Engineering,
        University of California, Davis, CA 95616, USA}
\thanks{$^*$Corresponding author, e-mail: {\tt\small lxflin@ucdavis.edu}}
}

\begin{document}

\maketitle
\thispagestyle{empty}
\pagestyle{empty}

\begin{abstract}

Multirotor unmanned aerial vehicle is a prevailing type of aerial robots with wide real-world applications.
The energy efficiency of the robot is a critical aspect of its performance, determining the range and duration of missions that can be performed. 
This paper studies the energy-optimal planning of the multirotor, which aims at finding the optimal ordering of waypoints with the minimum energy consumption for missions in 3D space.
The study is performed based on a previously developed model capturing first-principle energy dynamics of the multirotor.
We found that in majority of the cases (up to $95\%$) the solutions of the energy-optimal planning are different from those of the traditional traveling salesman problem which minimizes the total distance.
The difference can be as high as 14.9\%, with the average at 1.6\%-3.3\% and 90th percentile at 3.7\%-6.5\% 
depending on the range and number of waypoints in the mission. 
We then identified and explained the key features of the minimum-energy order by correlating to the underlying flight energy dynamics.
It is shown that instead of minimizing the distance, coordination of vertical and horizontal motion to promote aerodynamic efficiency is the key to optimizing energy consumption.

\end{abstract}

\section{Introduction}
Multirotor unmanned aerial vehicles (UAVs) have been seeing increasing application in a wide range of tasks, such as aerial photographing and inspection, package delivery, and intelligence, surveillance, and reconnaissance (ISR) among many others \cite{scaramuzza2022, karydis2017energetics}.
However, energy performance poses a significant challenge, limiting the endurance of widely used multirotor UAVs to approximately 30 minutes \cite{scaramuzza2022}.
The endurance is a major weakness compared with other mobile robots and presents a substantial obstacle for future applications.
In addition to the required advancements in high density energy storage technology, bridging this gap will require research on enhancing the energy efficiency of multirotor flight through planning and control.
Various levels of works have been performed in this broad research area, including modeling and prediction of energy consumption \cite{michel2022performance, scaramuzza2022, liu2017power}, 
energy-optimal control of flight trajectory \cite{Michel2023asme, morbidi2021practical}, mission planning \cite{bouzid2017quadrotor, wei2018coverage}, and swarm coordination \cite{mathew2015, modares2017ub}.

The focus of this work is motion planning at the mission level, with the goal of identifying the optimal ordering of waypoints in 3D space yielding the minimum energy consumption.
Existing research on this topic dates back to the classic Traveling Salesman Problem (TSP), 
which aims at minimizing the total distance of covering a series of waypoints \cite{oberlin2011, xu2019}.
Intuitively, it is reasonable to equate minimal distance with minimal energy consumption, 
under the notion (assumption) that energy use is simply proportional to the distance traveled. 
For example, in \cite{wei2018coverage, sathyan2015comparison, bouzid2017quadrotor}, the waypoint coverage plan is optimized for a single UAV 
with the goal of finding the shortest path;
in \cite{mathew2015, xie2022multiregional, modares2017ub}, similar works are performed for multiple UAVs or UAV swarms;
in addition, \cite{yu2019coverage} and others study the optimal route planning for the joint operation of a UAV and a ground vehicle for environment coverage.
However, is it really true that minimum distance always means minimum energy consumption?
Especially, what if the waypoints are in the more complicated 3D environment, which involves both horizontal and vertical motion? 
In reality, energy consumption is more complicated than just distance, as 
UAV motion could have major impacts on the aerodynamic efficiency of the rotor due to change in propeller inflow \cite{michel2022performance, Michel2023asme}.  
Therefore, different combinations of motion to cover the same set of waypoints could result in different energy consumption. 
There have been some works in literature aiming at minimizing the energy consumption instead of just the distance for path and/or mission planning \cite{dorling2016vehicle, morbidi2016minimum, schacht2018path}.
However, they are subject to major oversimplications of the critical UAV energy dynamics, including treating power consumption among hovering, horizontal, and vertical flights as equal, or assuming a constant rotor thrust/torque and angular velocity relationship (hence neglecting the impact of motion on rotor efficiency).

To address the gap in state of the art, this paper studies the energy-optimal waypoint-based mission planning, 
building on our previous modeling and control works. 
In \cite{michel2019multiphysical, michel2022performance}, a system-level model of the multi-physical dynamics of an octorotor vehicle, including rotor aerodynamics, motor and electronic speed controller (ESC) electro-mechanical dynamics, battery electrical dynamics, and airframe rigid-body dynamics, was formulated, parameterized, and validated.
The model was then used to develop a framework for energy-optimal vehicle path generation and following between two arbitrary waypoints in 3D space in \cite{michel2020optimal, michel2022acc, Michel2023asme}, 
demonstrating significant improvement over a baseline control approach.
In this work, we extend the optimization to the mission level, with the goal of finding the optimal ordering of multiple waypoints in 3D space with the minimum total energy consumption.
We first formulate the waypoint-based mission planning problem, where the (optimal) energy consumption between any two waypoints can be computed using our previous modeling and path generation works.
Optimization is then performed over a large number of missions with randomized waypoint locations. 
The energy consumption of the minimum-energy order is evaluated and compared with that of the minimum-distance order.
These results are discussed to demonstrate the energy savings achieved by the minimum-energy ordering approach under different conditions, including varying mission ranges and numbers of waypoints.
Detailed analysis of two sample missions are performed to identify and explain the features of the minimum-energy order by correlating to the underlying energy dynamics, giving valuable insights on the heuristics of energy-optimal mission planning.
Finally, one of these sample missions is tested in real-world operation.
Theoretical and experimental energy performance are shown to match well, validating the minimum-energy ordering results and demonstrating key behaviors related to energy performance observed and discussed in this work.



\section{Modeling of Multirotor Energy Dynamics}
To accurately capture the comprehensive physical dynamics of a multirotor, a system-level model is employed \cite{michel2022performance}. 
This model considers the aerodynamics of the propeller-rotor assembly \cite{textbook}, the electro-mechanical dynamics of the motor and electronic speed controller (ESC) \cite{lawrence2005efficiency}, the electrical dynamics of the battery \cite{lin2014lumped}, and the rigid body dynamics of the vehicle \cite{Powers2015Quadrotor}.
\begin{figure}[t]
	\centering
		\includegraphics[width=0.4\textwidth]{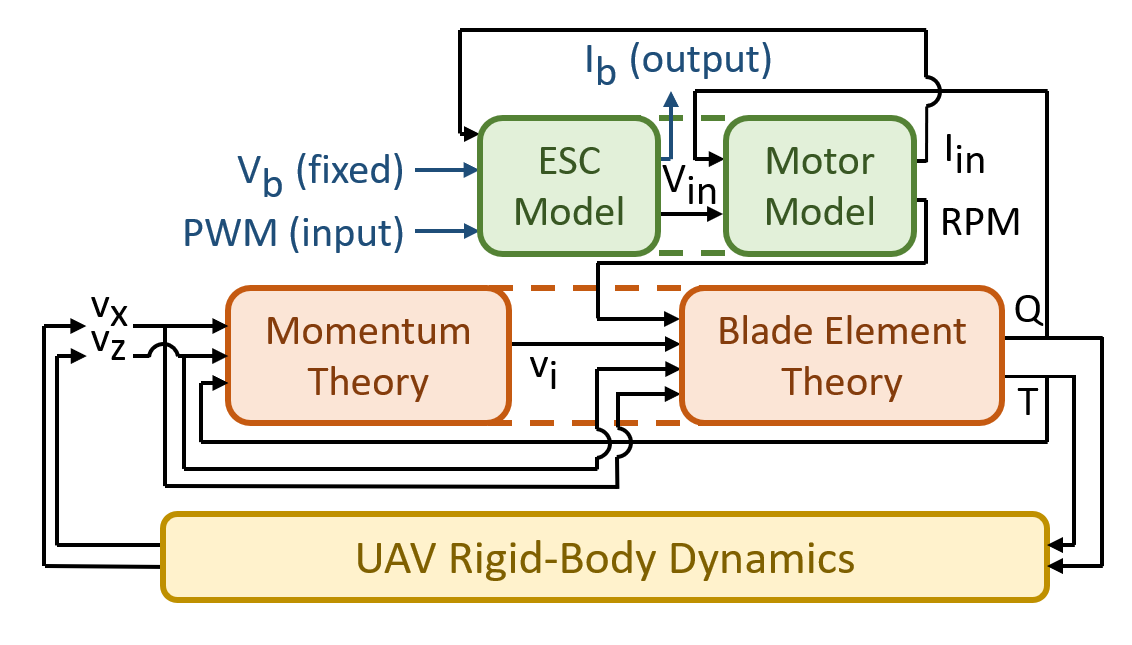}
	\caption{Block diagram of System-Level Model Architecture}
	\label{fig:ModelArchitecture}
\end{figure}
The integration of the subsystem dynamics into the overall system model is described in Fig. \ref{fig:ModelArchitecture}. 
The low-level inputs to the system are the Pulse-Width Modulation (PWM) commands to each motor. These commands instruct the Electronic Speed Controller (ESC) to regulate the motor input voltage $V_{in}$ as a fraction of the battery voltage $V_b$.
In response to the PWM commands, the motor draws current $I_{in}$ and rotates at a certain RPM, which is influenced by the torque load exerted by the rotor.
The rotation of the motor drives the rotors, with each rotor generating torque $Q_j$ and thrust $T_j$ based on the blade element theory.
These thrust and torque values are utilized in the rigid body dynamics model to calculate the motion of the multirotor, including each rotor's planar and perpendicular airspeed components ($v_{x,j}$ and $v_{z,j}$, respectively), which consequently affect rotor performance according to both the blade element and momentum theories.
The torque is also fed back to the motor, determining the motor speed and current,
which are then looped back to the ESC, to determine the current $I_b$ drawn from the battery. 
The current draw is an output of the system and can be multiplied by the battery voltage to calculate the power consumed during the operation.

In following subsections, we will briefly discuss the propeller aerodynamics and rigid body kinetics, which have key impacts on the energy-optimal mission planning studied in this work. 
Detailed information regarding the derivation and parameterization of the subsystem and overall models can be found in \cite{michel2022performance}.

\subsection{Key Subsystem Dynamics}
\subsubsection{Rigid Body Kinetics}
The motion of the vehicle is modeled based on the rigid body dynamics of the airframe. 
In this work, the following equations are considered, which are sufficient to describe the motion between two waypoints in 3D space, 
\begin{equation}
\begin{aligned}
    \ddot{X}&=\sum\limits_{j=1}^{N_p}T_{j}sin(\Theta)/m-C_{BD}\dot{X}|\dot{X}|/m \\
    \ddot{Z}&=\sum\limits_{j=1}^{N_p}T_{j}cos(\Theta)/m-g \\
    \ddot{\Theta}&=\tau_\Theta/J_\Theta 
    =\sum_{j=1}^{N_p} L_{\Theta,j} T_j /J_\Theta,    
\end{aligned}
\label{eq:2DMotion}
\end{equation}
which are solved over time to obtain the velocity, position, and orientation of the UAV.
Specifically, $X$ and $Z$ are the horizontal and vertical positions of the vehicle center of mass in the global frame, $\dot X$ and $\dot Z$ are the corresponding velocities, and $\Theta$ is the pitch angle.
In addition, ${\Sigma}T_{j}$ is the sum of thrusts of all rotors computed by the propeller model to be discussed next (with each rotor indexed by subscript $_j$ and total number of rotors $N_p$), $C_{BD}\dot{X}|\dot{X}|$ is the body drag force with $C_{BD}$ as the body drag coefficient, $\tau_\Theta$ is the total pitch-axis torque on the vehicle generated by the thrusts, $J_\Theta$ is the moment of inertia about the pitch axis, and $L_{\Theta,j}$ is the arm length of each rotor thrust to the pitch axis.  
Notably, while body drag as a result of horizontal motion is considered in Eqn. \ref{eq:2DMotion},
vertical drag is not considered 
for two reasons.
First, the rotor disks account for a significantly larger proportion of the vehicle's cross-section area in the vertical direction than in the horizontal direction.
The impact of vehicle motion on the rotors is already included as part of the blade element momentum theory calculations discussed below, 
and is hence not considered again in the body drag.
Second, the body drag is proportional to the velocity squared, and in typical operation, the vertical velocity $\dot Z$ tends to be significantly lower than the forward velocity $\dot X$.
Consequently, body drag in the vertical direction is typically much smaller and can therefore be neglected.
Finally, the velocities of the center of each rotor in the vehicle frame can be calculated as
\begin{equation}
\begin{aligned}
    v_{x,j}=&\dot{X}\cos{\Theta}-\dot{Z}\sin{\Theta} \\
    v_{z,j}=&\dot{X}\sin{\Theta}+\dot{Z}\cos{\Theta}+\dot{\Theta}x_j.
\end{aligned}
\label{eq:RotorInflow}
\end{equation}


\subsubsection{Propeller Aerodynamics}
\begin{figure}[t]
	\centering
		\includegraphics[width=0.35\textwidth]{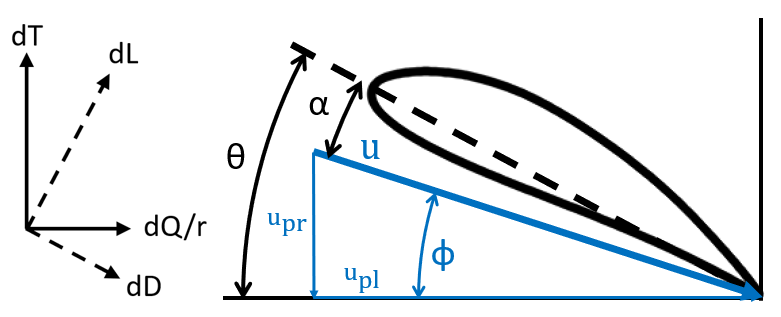}
	\caption{Geometries of a Blade Cross-section}
	\label{fig:CrossSection}
\end{figure}

The propeller model uses the blade element and momentum theory 
to calculate the thrust $T_j$ and torque $Q_j$ generated by each rotor, with propeller angular velocity $\omega$, and horizontal and vertical velocities $v_x$ and $v_z$ computed by the rigid body kinetics as inputs. 
Specifically, an infinitesimal segment of the rotor blade is shown in Fig. \ref{fig:CrossSection}, where the element thrust $dT$ and torque $dQ$ are, respectively, perpendicular to and parallel to the rotor plane. They are generated by the lift and drag forces $dL$ and $dD$ as the blade rotates, moving the illustrated cross-section towards the left of the figure.
The thrust and torque of the propeller blade can be computed by integrating $dT$ and $dQ$ along the blade length coordinate $r$ and averaged over one rotation cycle, 
\begin{equation}
\begin{aligned}
	&T_j=\int_{0}^{2\pi}\int_{R_0}^{0.97R}dTd{\psi}dr/2\pi \\
    & = \int_{0}^{2\pi}\int_{R_0}^{0.97R}({0.5 N\rho{u_{pl}}^2ca}(\theta-u_{pr}/u_{pl}))d{\psi}dr/2\pi \\
	&Q_j= \int_{0}^{2\pi}\int_{R_0}^{R}dQ d{\psi}dr/2\pi \\
    & = \int_{0}^{2\pi}\int_{R_0}^{R}{(0.5Nr\rho{u_{pl}}^2c}(\phi{a}(\theta-u_{pr}/u_{pl})+c_d))d{\psi}dr/2\pi, 
\end{aligned}
\label{eq:TQsaa}
\end{equation}
where 
$\psi$ represents the angular position of the blade along its rotating direction, $N$ is the number of blades in each rotor,
$\rho$ is the air density, $c_d$ is the propeller drag coefficient, $a$ is the lift coefficient factor, $c$ is the blade chord length,
and $\theta$ is the twist angle of the blade. 
The integration is performed from the base of the blade $R_0$ to 97\% of the tip $R$ instead of 100\% to approximate tip loss. 
In addition, the planar inflow velocity $u_{pl}$, perpendicular inflow velocity $u_{pr}$, and inflow angle $\phi$ can be computed as
\begin{equation}
\begin{aligned}
	&u_{pl}(r,\psi)=\omega r+v_{x} sin(\psi), \quad 
	u_{pr}(r)=v_i+v_{z} \\
	&\phi(r,\psi)={\tan}^{-1}{(u_{pr}/u_{pl})}, 
\end{aligned}
\label{eq:BladeElement}
\end{equation}
where $v_i$ is the propeller induced velocity that needs to computed by coupling with the momentum theory. 
These equations are also simplified using the small-angle approximation and the conditions $u_{pr}{\ll}u_{pl}$ and $dD{\ll}dL$, which have been validated in \cite{michel2019multiphysical}.

\subsection{Model-Based Energy Efficiency Analysis}
\label{ModelPerformance}
Based on the model, key factors with significant impacts on energy efficiency can be identified, which will enlighten the analysis on energy-optimal mission planning in the following sections.
First, it can be shown that rotor efficiency, defined as thrust generated per unit power input $T/P$, decreases as $\omega$ increases.
According to Eqn. (\ref{eq:TQsaa}), both $T$ and $Q$ are approximately proportional to $u^2_{pl}$, and $u_{pl}$ is approximately proportional to $\omega$ according to Eqn. (\ref{eq:BladeElement}). 
Consequently, the mechanical power input to the rotor ($P = Q\omega$) increases with $\omega$ more rapidly than thrust (cubic versus quadratic).
This effect is especially pronounced at high $\omega$, where a small increase in thrust demand can cause a disproportionately large increase in power consumption, hence reducing energy efficiency.
Second, the effect of vehicle-frame horizontal velocity $v_x$ and vertical velocity $v_z$ on rotor efficiency can be derived.
Increasing $v_z$ generally decreases rotor efficiency, 
as higher $v_z$ leads to higher $u_{pr}$ in Eqn. (\ref{eq:BladeElement}), and hence 
reduces the thrust in Eqn. (\ref{eq:TQsaa}). 
Meanwhile, increasing $v_x$ generally has a positive effect on rotor efficiency by increasing 
$u_{pl}$ as in Eqn. (\ref{eq:BladeElement}), and hence the thrust in Eqn. (\ref{eq:TQsaa}). 

\section{Problem Formulation for Mission Planning}
\label{Ch: Mission_Planning}

A mission considered in this work consists of a set of $N_W$ waypoints in 3D space, i.e. $\{W_i\}_{i=1}^{N_W}$, which the vehicle visits each once before returning to its starting position (origin $O$). 
The goal of mission planning is to minimize the total energy consumption over the whole mission,
\begin{equation}
    \min_{S} E = \sum_i^{N_W-1} E_{W_{s_i} \to W_{s_{i+1}}} + E_{O \to W_{s_{1}}} + E_{W_{s_{N_W}}\to O} 
\label{eq:PowCost}
\end{equation}
where the optimization variable $S = [s_1,s_2,...s_{N_W}]$ is the order of visiting the $N_W$ waypoints,
$E_{W_{s_i} \to W_{s_{i+1}}}$ denotes the energy cost for the vehicle to move between adjacent waypoints,
and $E_{O \to W_{s_{1}}}$ and $E_{W_{s_{N_W}}\to O}$ are the energy costs to start from and return to the origin.
The optimization is subject to constraints on UAV dynamics, including limits on acceleration, velocity, and pitch angle among others. 
In addition, the vehicle 
needs to stop and hover at each waypoint before proceeding to the next one. 
This mission structure 
is a realistic setting with wide real world applications, including surveillance, inspection, and delivery tasks, in which a multirotor UAV needs to stop and take images, perform measurements, or drop packages at each waypoint.

\begin{figure}[b]
	\centering
		\includegraphics[width=0.4\textwidth]{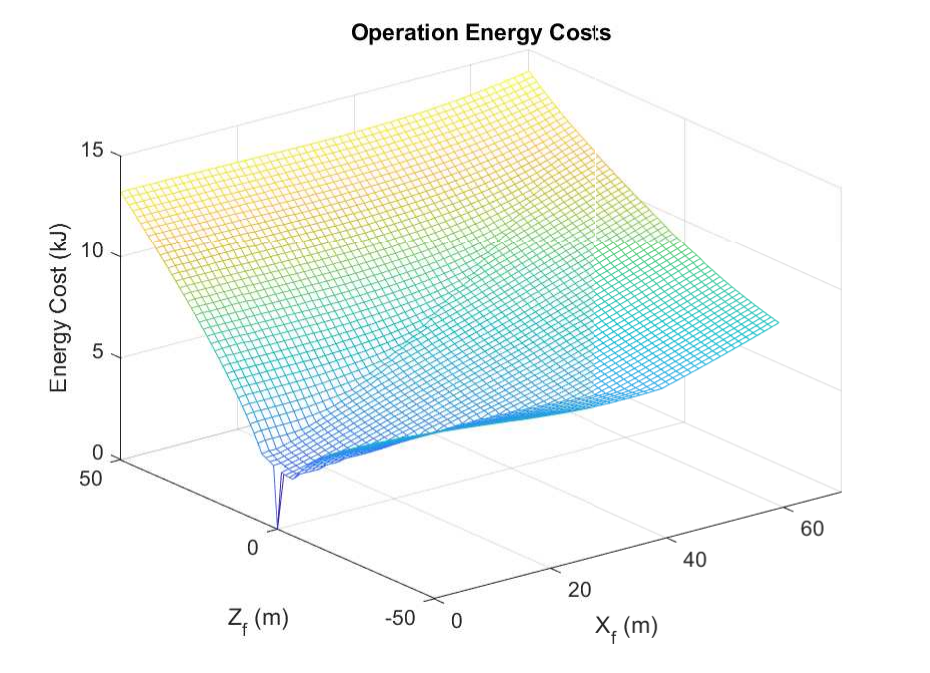}
	\caption{Reference Table for Optimal Energy Costs between Waypoints over a Range of Horizontal and Vertical Displacements} 
	\label{fig:EnergyCostTable}
\end{figure}

The energy cost $E$ in Eqn. (\ref{eq:PowCost}) is evaluated based on the previously introduced multirotor model.
In our previous work \cite{michel2020optimal}, a method was developed to obtain the energy-optimal trajectory between arbitrary waypoints in 3D space along with the optimal energy cost.
To facilitate the computation, polynomial approximations of optimized trajectories were further developed in \cite{michel2022performance, Michel2023asme}.  
The polynomial approximations were fit using optimized trajectories generated over a range of operations, 
and, when evaluated in real-time control, demonstrated good agreement with the optimized trajectories. 
Furthermore, a reference table of the optimal energy costs using the polynomial trajectories was pre-calculated for operations between waypoints over a range of horizontal ($X_f$) and vertical ($Z_f$) displacements,  
as plotted in Fig. \ref{fig:EnergyCostTable}. 
In this work, interpolation from this table is used to predict the energy cost between any pair of waypoints in Eqn. (\ref{eq:PowCost}). 

In this work, we use different methods to find the optimal orders for large number of randomized missions with number of waypoints $N_W$ between 6 and 10.
Noted that the problem is essentially a variant of the NP-hard TSP problem with the cost of travelling between nodes defined by the energy map, and solutions can be computationally difficult to find for large $N_W$.
The methods we used include exhaustive search, which gives the exact solutions, and Mixed Integer Linear Programming (MILP), which is an efficient approach for solving the NP-hard TSP problem. 
The solutions given by the two methods have been verified to be the same.  


To demonstrate the performance improvement achieved by the energy-optimal mission planning, the minimum-distance orders of the waypoints are also determined as the baseline for comparison,   
which can be more easily found without understanding the multirotor energy dynamics. 
Specifically, we consider 3 types of minimum-distance orders, namely those minimizing the horizontal distance $X_M$, vertical distance $Z_M$, and total distance $D_M$. 
The energy consumption of these orders are evaluated the same way as the energy-optimal order. 
It will be seen that by analyzing and comparing the energy consumption against these 3 orders, the underlying mechanisms of the energy-optimal order can be better understood, which are heavily dependent on the energy dynamics of the horizontal and vertical motion.
Also noted that each minimum distance order can be traversed in two directions, 
which could have different energy implications.
Additionally, for any mission with $N_W>2$, multiple minimum vertical distance orders could exist and we pick the one with the shortest horizontal distance $X_M$ as the minimum order for analysis.

\section{Result Analysis}
\label{MissionEnergyPerformance}



To evaluate the energy performance of mission planning, 
large numbers of missions ($500-5000$) with random waypoint locations are generated over a series of $N_W$ and operation ranges.
For each of these missions, the min-energy order and the min-distance orders are identified, and their energy costs are compared.
Two sample missions will be discussed in detail to demonstrate the key features of the minimum energy order related to the fundamental energy dynamics of the multirotor.
The vehicle considered in this work is an octorotor, of which the model and trajectory generation have been studied and validated in our 
previous works \cite{michel2022performance, Michel2023asme}. 

\subsection{Energy Performance of Different Orders}
\begin{figure}[t]
	\centering
		\includegraphics[width=0.38\textwidth]{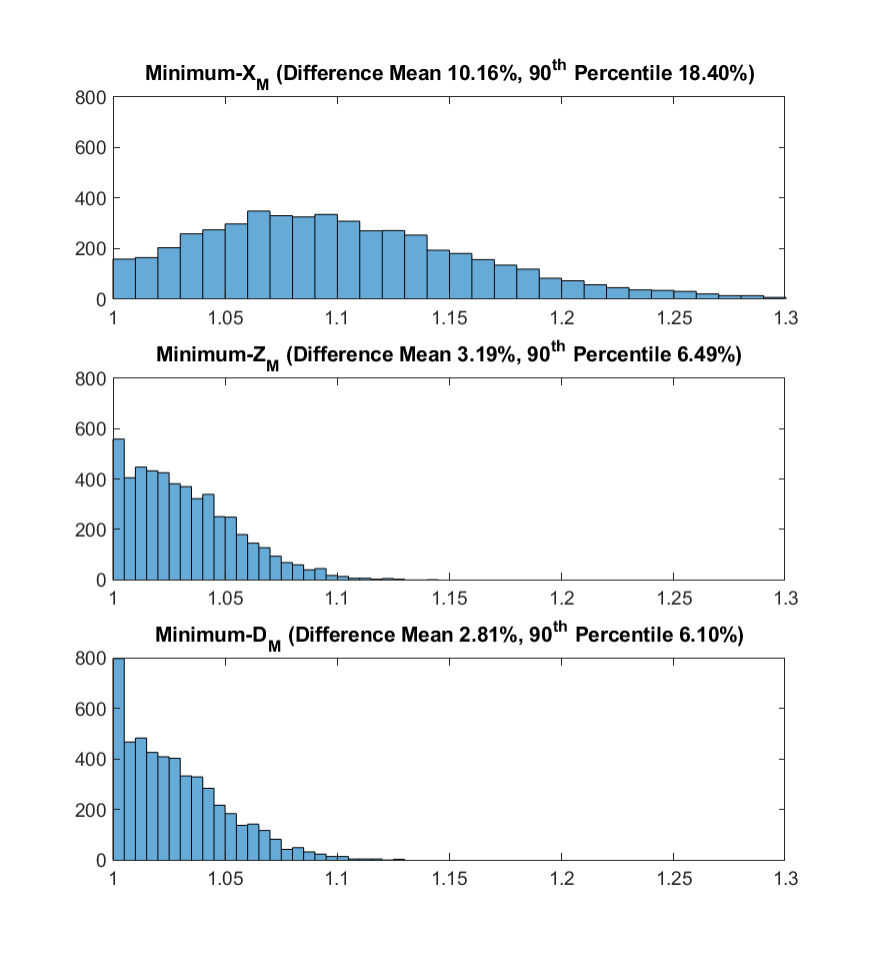}
	\caption{Histograms of Ratio of Energy Cost of Baseline Orders (Min-$X_M$, $Z_M$, and $D_M$) over Min-Energy Order for 5000 Randomized Missions, with $N_W=8$ and Ranges of $-30m<X<30m$, $-30m<Y<30m$, $-25m<Z<25m$}
	\label{fig:PredictionHistograms}
\end{figure}

The results comparing the energy performance of the different ordering approaches are presented in Table \ref{table:MissionPredictionResults}
which shows the increase in energy consumption of the 3 min-distance orders over the min-energy order in percentage. 
The histogram for one combination of mission parameters (8 waypoints within range $-30m<X<30m$, $-30m<Y<30m$, $-25m<Z<25m$) are given in Fig. \ref{fig:PredictionHistograms}, where the x-axis denotes the ratio of energy cost of each min-distance order to that of the min-energy order. 
It is seen that for majority of the missions, the minimum-distance order does not give the minimum energy consumption, 
as indicated by the "\% Min-$D_M$ $\neq$ Min-Energy" column in Table \ref{table:MissionPredictionResults}, 
e.g. in 83.3\% of the missions for $N_W = 6$ and 95.8\% for $N_W = 10$.
The maximum difference between the two orders is as high as 10.6-14.9\%
depending on the mission ranges and number of waypoints,
with the average and 90$^\text{th}$ percentile difference at 1.6\%-3.3\% and 3.7\%-6.5\% respectively.
These results indicate the significant energy loss when planning the mission just based on distance.
Meanwhile, the difference is more prominent when comparing the minimum-energy order with minimum-$X_M$ and minimum $Z_M$ orders.
Specifically, minimizing $Z_M$ results in average increase in energy consumption of 2.7\%-5.1\%, while minimizing $X_M$ results in average increase of 4.2\%-16.6\%.
Regarding the 3 baseline minimum distance ordering, although minimizing the total distance $D_M$ uses the least energy on average, the performance is not consistent across individual missions.
Specifically, minimizing $Z_M$ is more efficient in between 34.8\% and 45.6\% of missions, and minimizing $X_M$ is more efficient in between 1.6\% and 5.3\% of missions. 
\begin{table*}[t]
    \centering
    \begin{tabular}{|c|c|c|c||c|c||c|c||c|c|c|c|}
        \cline{5-11} \multicolumn{4}{c|}{} & \multicolumn{7}{c|}{Energy Cost Relative to Min-Energy Order} \\
        \cline{2-12} \multicolumn{1}{c}{} & \multicolumn{2}{|c|}{Range (m)} & \multirow{2}{*}{\begin{tabular}[b]{@{}c@{}}No. of\\Missions\end{tabular}} & \multicolumn{2}{c||}{Min-$X_M$} & \multicolumn{2}{c||}{Min-$Z_M$} & \multicolumn{3}{c|}{Min-$D_M$} & \multirow{2}{*}{\begin{tabular}[b]{@{}c@{}}\% Min-$D_M$ \\ $\neq$ Min-Energy\end{tabular}}\\ \cline{1-3} \cline{5-11} $N_W$ & X\&Y & Z & & Mean & 90$^\text{th}$ {\%}ile & Mean & 90$^\text{th}$ {\%}ile & Mean & 90$^\text{th}$ {\%}ile & Max &\\ \hline \hline
        6 & $\pm$30 & $\pm$25 & 5000 &  +7.16\% & +14.58\% & +2.72\% & +5.99\% & +2.15\% & +5.19\% & +14.87\% & 83.32\%\\ \hline
        8 & $\pm$20 & $\pm$25 & 5000 & +16.25\% & +26.68\% & +2.53\% & +5.18\% & +2.62\% & +5.75\% & +13.51\% & 93.70\%\\ \hline
        8 & $\pm$30 & $\pm$25 & 5000 & +10.16\% & +18.40\% & +3.19\% & +6.49\% & +2.81\% & +6.10\% & +12.93\% & 91.66\%\\ \hline
        8 & $\pm$40 & $\pm$25 & 5000 &  +6.12\% & +12.66\% & +4.29\% & +8.75\% & +2.12\% & +4.87\% & +12.60\% & 85.72\%\\ \hline
        8 & $\pm$30 & $\pm$15 & 5000 &  +4.27\% &  +9.76\% & +4.98\% & +9.29\% & +1.58\% & +3.76\% & +10.58\% & 84.80\%\\ \hline
        10 & $\pm$30 & $\pm$25 & 500 & +12.57\% & +20.18\% & +3.69\% & +7.08\% & +3.36\% & +6.55\% & +13.27\% & 95.80\%\\ \hline
    \end{tabular}
    \caption{Average, Maximum, and 90$^\text{th}$ Percentile Energy Cost Increase over Min-Energy Order of Min-$X_M$, $Z_M$, and $D_M$ Orders for Missions with Varying Number of Waypoints and Ranges}
    \label{table:MissionPredictionResults}
\end{table*}

The key to interpret these results is to account for the stronger impact of vertical motion on energy performance compared with horizontal motion,
which will be discussed with details in the next subsection. 

\subsection{Analysis of Sample Missions}
\label{SampleMission}
\begin{figure}[b]
	\centering
		\includegraphics[width=0.45\textwidth]{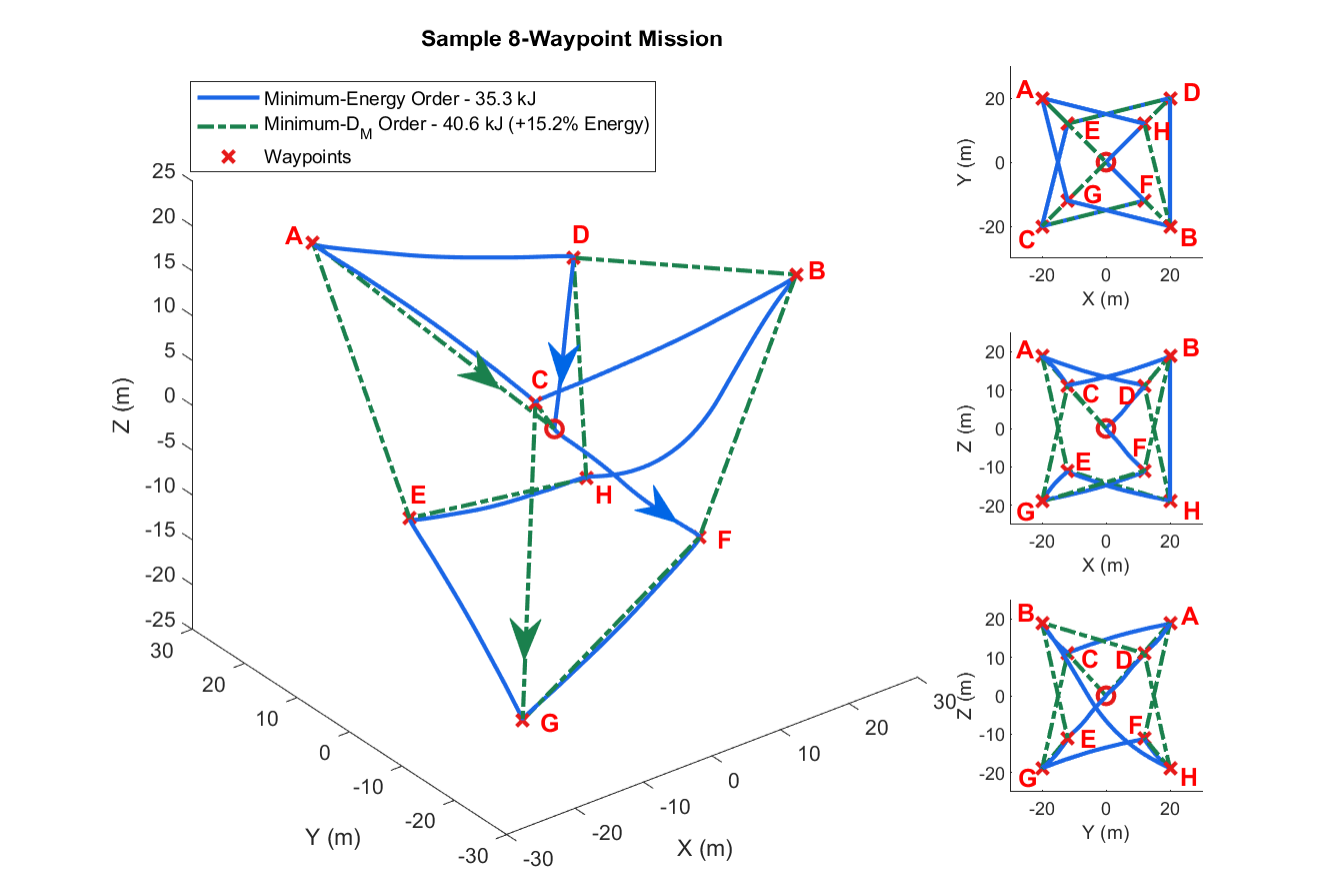}
	\caption{Minimum-Energy (Solid) and Minimum-Distance (Dashed) Orders for 8-Waypoint Sample Mission, Including Multiple Viewing Angles.}
	\label{fig:SampleMission}
\end{figure}

To further explore and explain key features of the minimum-energy ordering, two sample missions 
are chosen for discussion.
The waypoints of the first mission are specified in Table \ref{table:SampleWaypoints}, along with the minimum-energy and minimum-distance orders shown in Fig. \ref{fig:SampleMission}.
Simulation studies are then performed to evaluate the energy performance of different ordering approaches.

\begin{table}[b]
    \centering
    \begin{tabular}{|r|cccccccc|}
    \cline{2-9} \multicolumn{1}{r|}{} & A & B & C & D & E & F & G & H \\
    \hline X & -20 & 20 & -12 & 12 & -12 & 12 & -20 & 20 \\
    Y & 20 & -20 & -12 & 12 & 12 & -12 & -20 & 20 \\
    Z & 19 & 19 & 11 & 11 & -11 & -11 & -19 & -19 \\ \hline
    \end{tabular}
    \caption{Waypoints of Sample Mission 1}
    \label{table:SampleWaypoints}
\end{table}

The energy performance of different ordering approaches 
is given in Table \ref{table:SampleMissionResults}.
It is seen that 
the min-$D_M$ order uses 15.2\% more energy than the energy-optimal order for this mission, even though the total distance traveled is 5\% shorter (284.4 m versus 299.3 m).
Meanwhile, the min-$X_M$ order uses 29.1\% more energy, and the min-$Z_M$ order uses either 11.7\% or 7.5\% more (depending on direction) than the min-energy order, but outperforms the min-$D_M$ order even though the distance traveled is 13\% longer (321.4 m versus 284.4 m).
\begin{table*}[t]
    \centering
    \begin{tabular}{|c||c|c|c|c|}
    \hline Order &  Min-Energy & Min-$D_M$ & Min-$X_M$ & Min-$Z_M$ (2 directions)
         \\ \hline \hline
        Energy Use (kJ) & 35.3 & 40.6 & 45.5 & 39.4/37.9  \\ 
       \% Increase & - & 15.2 & 29.1 & 11.7/7.5 \\ \hline
        \multicolumn{1}{|c||}{$D_M$ (m)} & 299.3 & 284.4 & 311.6 & 321.4 \\ \hline
        \multicolumn{1}{|c||}{$X_M$ (m)} & 271.9 & 189.5 & 183.2 & 291.3 \\ \hline
        \multicolumn{1}{|c||}{$Z_M$ (m)} & 108 & 174 & 240 & 76 \\ \hline
    \end{tabular}
    \caption{Energy and Distance Metrics of 8-waypoint Sample Mission under Different Ordering Approaches}
    \label{table:SampleMissionResults}
\end{table*}

Key features of each ordering approach can be analyzed based on this sample mission. 
First, since vertical motion generally has a stronger (negative) impact on energy consumption than horizontal motion as discussed in section \ref{ModelPerformance}, the min-energy order tends to have larger $X_M$ (horizontal distance) but lower $Z_M$ (vertical distance) than the min-$D_M$ order, and the min-$X_M$ order performs very poorly.
It is seen from Fig. \ref{fig:SampleMission} that under the min-energy order, the vehicle 
visits all waypoints on the lower half plane first (Waypoints F, G, E, H), and then proceeds to the top half to minimize the vertical operations. 
By contrast, the min-$D_M$ order moves up and down repeatedly, and hence contains multiple vertical operations with high energy costs.
Second, the min-energy order is also not identical to the min-$Z_M$ order. 
In fact, the difference in $Z_M$ between the two orders (22 m) is greater than the difference in $X_M$ (19.4 m).
This behavior reflects the complexity of the tradeoffs between horizontal and vertical motion, and the need for the physics-based energy consumption computation and optimization to balance the two motions across the waypoints.
Specifically, long operations along one axis, especially the vertical axis, are generally less efficient than diagonal operations with motion along both axes.
For example, 
the vehicle consumes 8.2 kJ to move 50 m along X-axis while climbing 30 m along Z-axis, compared to 7.3 kJ for climbing 30 m alone and 5.2 kJ for moving 50 m forward alone. 
It is seen that adding horizontal motion to climbing barely increases the energy consumption, while significantly increasing the distance traveled and hence improves the overall energy efficiency.
This is because moving long distance along the Z-axis requires a long operation time due to the typical lower vertical velocity limit. 
Therefore, motion along the other (horizontal) axis can be completed simultaneously under relatively low velocity with minimal additional thrust, and hence only slightly increases the energy consumption.
In addition, a low horizontal velocity can also increase rotor efficiency due to the aerodynamic effects, as discussed in section \ref{ModelPerformance}, partially offsetting the increased thrust required for diagonal motion.
Therefore, in this mission, the min-energy order is mostly the same as the min-$Z_M$ order, except that the former commands the vehicle to climb diagonally from point H to point B, while the latter moves vertically to D and then horizontally to B.
The energy saving in this motion sequence accounts for most of the advantage of the min-energy order over the min-$Z_M$ order.

\begin{figure}[b]
	\centering
		\includegraphics[width=0.45\textwidth]{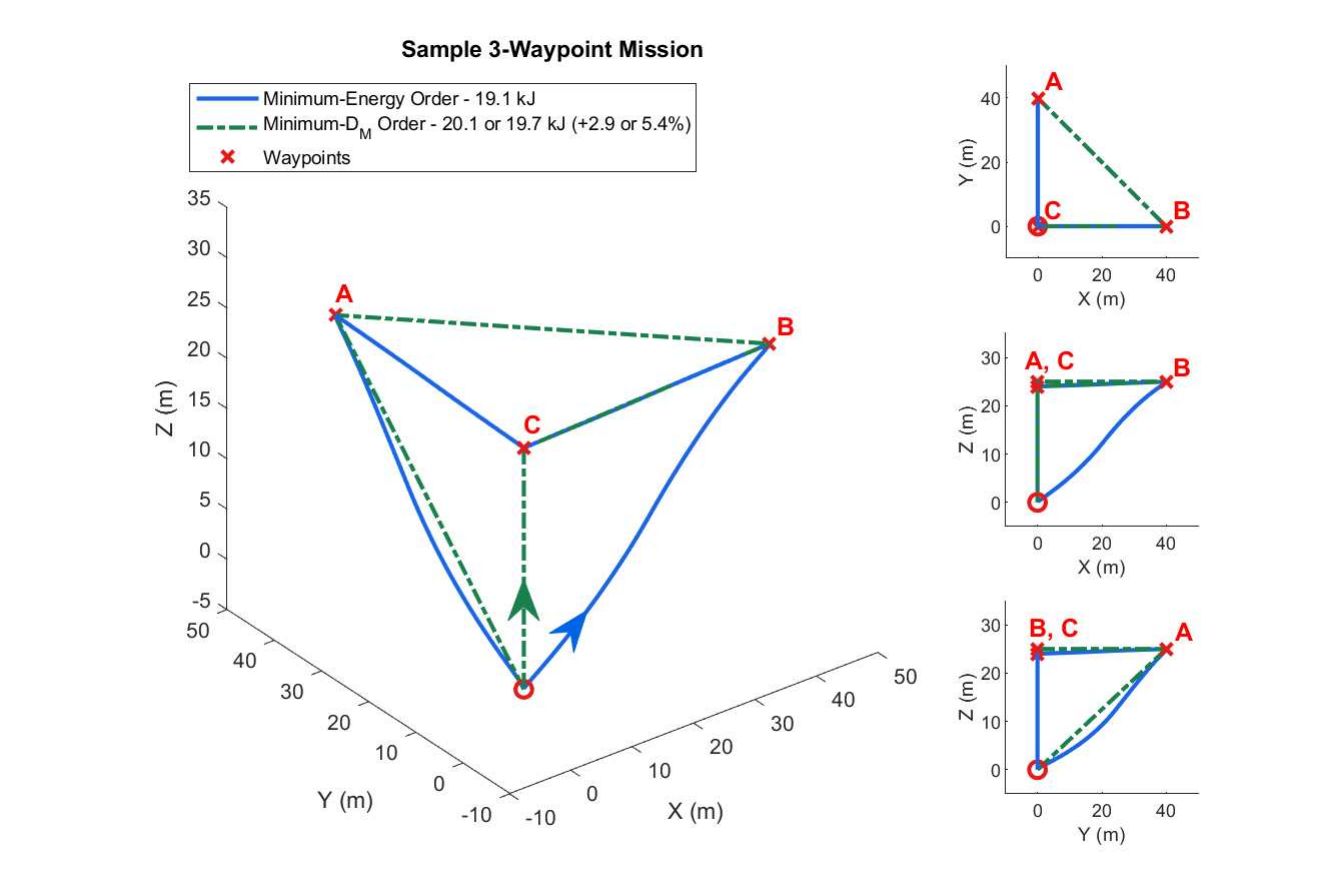}
	\caption{Minimum-Energy (Solid) and Minimum-Distance (Dashed) Orders of 3-Waypoint Sample Mission, Including Multiple Viewing Angles.}
	\label{fig:SampleMission3WP}
\end{figure}

For further analysis, a simpler 3-waypoint mission, consisting of the waypoints A, B, and C at $(0,40,25)$,  $(40,0,25)$, and $(0,0,24)$, is shown in Fig. \ref{fig:SampleMission3WP} for discussion.
In this mission, the min-$D_M$ order is the same as the min-$X_M$ and $Z_M$ orders, and, depending on direction of travel, uses either 19.7 or 20.1 kJ of energy, which is 2.9 or 5.4\% more than the 19.1 kJ used by the min-energy order.
Interestingly, $D_M$, $X_M$, and $Z_M$ of the min-$D_M$ order, which are 167.8 m, 136.6 m, and 50 m respectively, are all shorter than those of the min-energy order, which are 174.4, 160, and 52 m respectively. 
This result, though counterintuitive, is primarily due to the fact that the min-energy order prioritizes diagonal flight with both horizontal and vertical motion.
In the min-$D_M$ order, the vehicle climbs purely vertically from the origin to point C, using 6.22 kJ.
In the min-energy order instead, it climbs diagonally to point B, using only 0.83 kJ more energy while covering a significant horizontal distance.
The vehicle then travels to points C and A in two short horizontal operations before returning to the origin, while the min-$D_M$ order replaces the motion from C to A with a longer operation from B to A.
This longer operation by the min-$D_M$ order uses 1.39 kJ more energy, significantly outweighing the energy saved in the previous climb for a net increase of 0.56 kJ (2.9\%) over the min-energy order.
Additionally, the other direction of this min-$D_M$ order uses 0.47 kJ (2.4\%) more than its reverse for a total increase of 1.02 kJ (5.4\%) relative to the minimum energy order.
However, as the direction of travel of a waypoint order has no effect on the distance travelled, the min-$D_M$ order is not able to determine which direction is more efficient, and hence reduce the energy usage.

\section{Experimental Validation}
\label{Sect:Experimental}
\begin{figure}[t]
	\centering
		\includegraphics[width=0.45\textwidth]{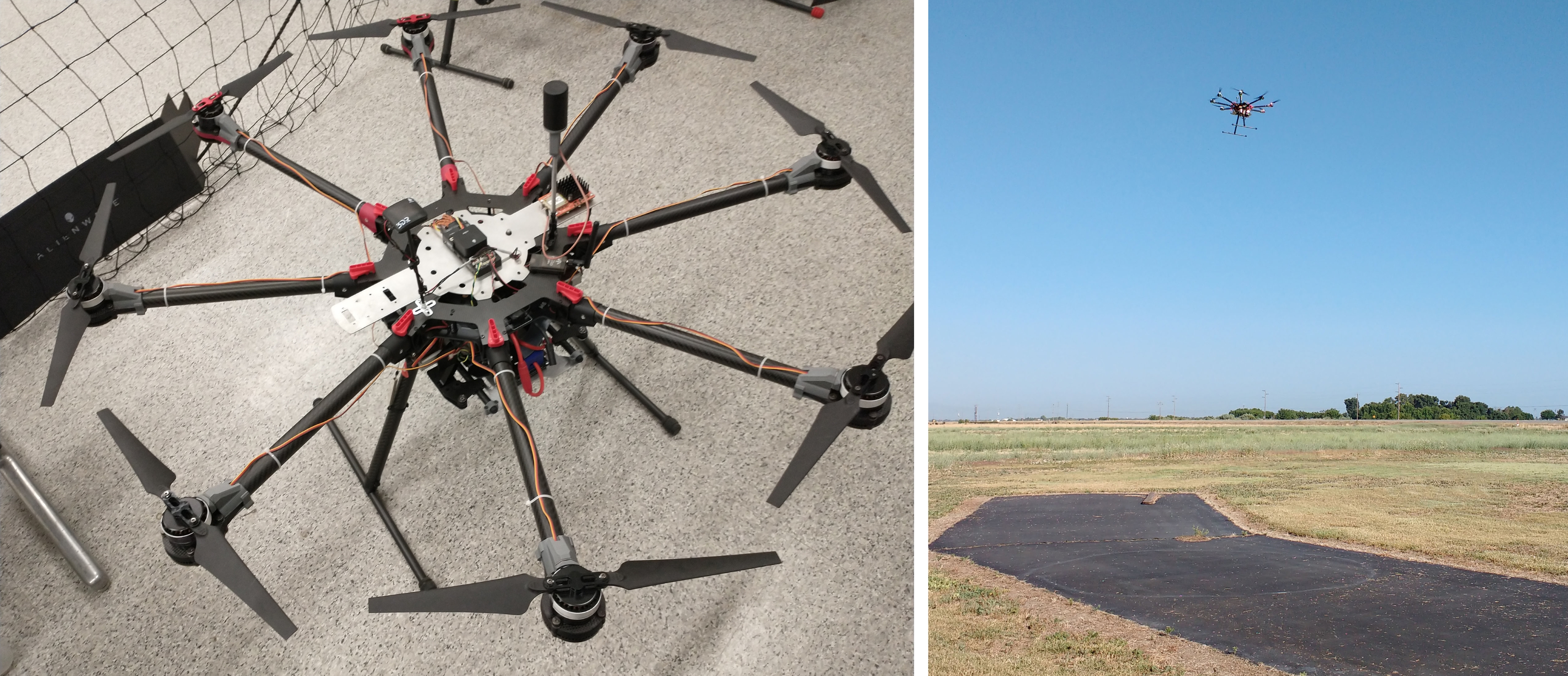}
	\caption{Left: Octorotor Testing Vehicle; Right: UAV Testing at Woodland-Davis Aeromodelers Test Site}
	\label{fig:UAV2}
\end{figure}

In this section, real-world flight test data are presented to validate the theoretical and simulation results. 
%
A test vehicle, shown in Fig. \ref{fig:UAV2}, was used for real-world experiments.
As described in \cite{michel2022performance}, this vehicle was built using the airframe and propulsion system (including propellers, motors, and ESC) of a DJI Spreading Wings S1000+ octorotor, and retrofitted with control and sensing systems including a Pixhawk 2.1 Flight Controller, a Piksi Multi GNSS receiver, a 3DR uBlox GPS with compass, a battery current transducer, and a voltage sensor.
The experiments were conducted at the Woodland-Davis Aeromodelers field, which is an open-air flat grass UAV test ground in Davis, CA, shown in Fig. \ref{fig:UAV2}.
A baseline controller, referred as the High-Velocity-Baseline (HVB) 
described in \cite{michel2022performance}, was also used for performance comparison. 

\subsection{Validation of 3-Waypoint Mission}
\label{Sub:Valid3WP}
We first present the experimental validation of the 3-waypoint sample mission previously discussed in Section \ref{SampleMission} and shown in Fig. \ref{fig:SampleMission3WP}. 
Each operation of the minimum-distance and minimum-energy orders was tested individually, and the energy consumption of each operation was added up to find the total energy used for the complete mission.
As in \cite{Michel2023asme}, operation energy cost was measured as the total energy drawn from the battery from the start of the operation to the moment the vehicle enters within 4 m of the operation endpoint.
This approach allows the vehicle to start each operation at a consistent and stable hover position, minimizing the impact of the previous operations and disturbances. 
In addition, each operation was repeated multiple times and the energy consumption was averaged to further mitigate the variations caused by disturbance (such as wind).

\begin{table}[h]
    \resizebox{0.45\textwidth}{!}{
    \centering
    \begin{tabular}{|c|c|c|c|c|c|c|c|c|c|c|}
        \cline{1-5} \cline{7-11} \multirow{2}{*}{\begin{tabular}[b]{@{}c@{}}Minimum-\\Energy\end{tabular}} & \multicolumn{4}{c|}{Operation \#} & \quad &
        \multirow{2}{*}{\begin{tabular}[b]{@{}c@{}}Minimum-\\Distance\end{tabular}} & \multicolumn{4}{c|}{Operation \#}\\
         \cline{2-5} \cline{8-11} & 1 & 2 & 3 & 4 & & \quad & 1 & 2 & 3 & 4 \\
         \cline{1-5} \cline{7-11} $X_f$ (m) & 40 & 40 & 40 & 40 & \quad & $X_f$ (m) & 40 & 56 & 40 & 0\\
         \cline{1-5} \cline{7-11} $Z_f$ (m) & 25 & -1 & 1 & -25 & \quad & $Z_f$ (m) & 25 & 0 & -1 & -24\\
         \cline{1-5} \cline{7-11}
    \end{tabular}}
    \caption{Operations Included in Minimum-Energy and Minimum-Distance Orders for 3-Waypoint Mission}
    \label{table:MissionOps}
\end{table}
The operations in the two orders are shown in Table \ref{table:MissionOps}.
As two operations are repeated between orders, only six unique operations are required to determine the total cost of both orders.
The results for these six operations are shown in Table \ref{table:MissionExpValSim},
including results for the Polynomial-Trajectory-Following (PTF, energy-optimal) and High-Velocity-Baseline (HVB) controllers, as well as the total energy costs for each pairing of the controller and order.
\begin{table}[t]
    \centering
    \resizebox{0.45\textwidth}{!}{
    \begin{tabular}{|c|cc|c||cc|c|}
        \cline{2-7} \multicolumn{1}{c|}{} & \multicolumn{6}{c|}{Operation Energy Cost (kJ):}\\
        \cline{2-7} 
        \multicolumn{1}{c|}{}& \multicolumn{3}{c||}{Experimental} & \multicolumn{3}{c|}{Simulated} \\
        \hline
         Operation: & PTF  & HVB  & \% Diff.: & PTF  & HVB  & \% Diff.:\\ \hline
         (40, 25)   & 6.57 & 7.58 & -13.36\%  & 7.04 & 8.10 & -13.01\% \\ 
         (40, 1)    & 3.65 & 4.17 & -12.40\%  & 4.52 & 5.67 & -20.31\% \\ 
         (40, -1)   & 3.67 & 4.23 & -13.17\%  & 4.36 & 5.53 & -21.19\% \\ 
         (40, -25)  & 2.95 & 4.69 & -37.22\%  & 3.19 & 4.77 & -33.12\% \\ 
         (56.6, 0)  & 4.76 & 5.11 & -6.80\%   & 5.74 & 6.62 & -13.19\% \\ 
         (0, -24)   & 2.53 & 5.30 & -52.30\%  & 2.99 & 4.97 & -39.81\% \\ \hline \multicolumn{7}{c}{\quad}\\ \cline{2-7} 
         \multicolumn{1}{c|}{} & \multicolumn{6}{c|}{Mission Energy Cost (kJ):} \\ 
        \cline{2-7} 
        \multicolumn{1}{c|}{}& \multicolumn{3}{c||}{Experimental} & \multicolumn{3}{c|}{Simulated} \\ \hline
         Order:    & PTF  & HVB  & \% Diff. & PTF  & HVB  & \% Diff. \\ \hline
         Min-E:    & 16.83 & 20.67 & -18.54\% & 19.11 & 24.07 & -20.59\% \\ 
         Min-$D_M$:& 17.52 & 22.21 & -21.11\% & 20.14 & 25.21 & -20.13\% \\ \hline
         \% Diff.:  & -3.92\% & -6.95\% & -24.20\% & -5.09\% & -4.54\% & -24.20\%\\ \hline
    \end{tabular}}
    \caption{Experimental and Simulation Results for Validation of 3-Waypoint Sample Mission}
    \label{table:MissionExpValSim}
\end{table}

Based on these results, using the PTF controller combined with the minimum-energy order results in a total mission energy cost of 16.83 kJ, while the same controller with the Min-$D_M$ order uses 17.5 kJ.
The min-E order therefore achieves an energy cost reduction of 3.92\%, capturing the majority of the 5.09\% reduction achieved in simulation.
Experimental conditions may explain the slight discrepancy in these results.
As discussed in \cite{Michel2023asme}, variable wind conditions on particular may have significant impacts on energy cost, and while the measures previously described were used to mitigate these impacts, such effects cannot be completely eliminated.
Meanwhile, the baseline HVB controller uses 20.67 kJ under the min-E order and 22.21 kJ under the min-$D_M$ order respectively.
The min-E order therefore reduces energy cost by 6.95\%, which is slightly larger than the result (4.54\%) in simulation, and experimental disturbances again likely account for some of this variation.
Finally, the combined PTF controller and min-E order are shown to achieve a total energy cost of 16.83 kJ, which is 24.20\% lower than the 22.21 kJ used by the combined HVB controller and Min-$D_M$ order and in excellent agreement with the simulated result.
This substantial energy cost reduction demonstrates the significant benefits of combining energy-optimal mission planning with the previously-developed energy-optimal trajectory planning and feedback control.

It is noted that there is an overall discrepancy between the energy costs in simulation and experiments, as was also observed in \cite{Michel2023asme}.
Specifically, the experimental flights use less energy (on average around 13\%), 
as summarized in Table \ref{table:MissionExpValSim}.
This discrepancy is likely due to the disturbances in experimental testing conditions (most notably wind) and the imperfection in modeling. 
Nevertheless, the discrepancy tends to have equal impacts on different control and ordering approaches, 
and hence the improvements achieved by energy-optimal ordering are similar and consistent in experiments and simulation, 
demonstrating its ability to improve vehicle energy performance in real-world operation.

\subsection{Validation of Diagonal Descent}
\label{Sub:ValidDiagonal}
According to our model-based theoretic and simulation analysis,  
one notable behavior observed is that adding a horizontal component to descending flight can allow significant additional distance to be covered with a comparatively small increase in energy cost.
For instance, for the three-waypoint sample mission, descending diagonally to the origin from point A uses only 16.58\% more energy but covers 88\% longer distance than descending vertically from point C, indicating significantly higher energy efficiency.
Furthermore, we have found that in some cases a combined descending and forward operation can even use less energy than a purely descending operation with the same vertical displacement.
To demonstrate this interesting behavior, three additional operations were tested, 
including one diagonal (forward) descent with displacements of 30 m along both $x$ and $z$ axis, one 30 m horizontal flight, and one 30 m descent flight, with at least three repeated trials for each operation.
The average energy cost of the diagonal descent (2.92 kJ) was found to be 12.4\% lower than that of the pure vertical flight (3.34 kJ), and 2.87\% lower than that of the pure horizontal (3.01 kJ) flight.
To understand this effect, recall that, as discussed in section \ref{ModelPerformance}, forward motion can improve rotor efficiency by increasing the propeller inflow.
During simultaneous descent and forward flight, even though additional thrust is needed for the forward motion, the increase in power consumption is minimal because the rotor thrust-to-power ratio is higher at lower angular velocities (typical of descent flight due to lower thrust demand), as discussed in section \ref{ModelPerformance}.
Consequently, the improvement in rotor efficiency outweighs the increase in thrust, resulting in a net decrease of energy consumption.

\section{Conclusions}
In this paper, energy-optimal planning of 3D waypoint-based UAV missions is studied by considering the underlying energy dynamics.
An optimization problem is set up to find the ordering of the waypoints with minimum energy consumption,
and optimization is performed over a large number of missions with randomized waypoint locations. 
It is found that 
the minimum-energy order is not identical to the minimum-distance order in majority of the cases, 
e.g. $>$ 95\% of 500 missions with 10 waypoints.
The difference in energy consumption between the two orders can be as high as 14.9\%, 
with the average at 1.6\%-3.3\% and 90th percentile at 3.7\%-6.5\% among missions of varying ranges and number of waypoints.
We then managed to identify several features of the min-energy order by comparing with the min-distance order.
First, it is important to minimize the number of vertical flights in a mission. 
Second, coupling vertical motion with horizontal motion could significantly promote the vehicle energy efficiency.
In the climbing case, simultaneous horizontal motion only slightly adds to the energy consumption while significantly increases the distance covered horizontally. 
In the descent case, the effect is even more prominent to the point that additional horizontal motion could actually reduce the energy consumption in some cases. 
These results were then explained by correlating to underlying fundamental UAV energy dynamics, 
especially the impact of motion on the aerodynamic efficiency. 
The insights generated by the work could greatly benefit the research on multirotor UAV energy optimization and range extension, enabling more areas of profound applications in aerial robotics and transportation.

\bibliographystyle{IEEEtran}
\bibliography{reference}     

\end{document}